\documentclass[letterpaper,10pt,conference]{ieeeconf}
\IEEEoverridecommandlockouts
\usepackage[T1]{fontenc}

\usepackage{xparse}
\usepackage{amsmath,amssymb,amsfonts} 
\usepackage{color}
\usepackage{graphicx}
\usepackage{calc}
\usepackage{cite}
\usepackage{url}
\usepackage{calc}
\usepackage{adjustbox}

\usepackage{tikz}
\usetikzlibrary{calc}
\definecolor{ply}{RGB}{222,184,135}
\definecolor{top_red}{RGB}{154,61,53}
\definecolor{base_blue}{RGB}{46,49,86}

\IEEEtriggeratref{18}

\usepackage{cleveref}
\crefformat{equation}{(#2#1#3)}
\Crefname{figure}{Fig.}{Figs.}
\crefname{figure}{Fig.}{Figs.}
\crefformat{figure}{#2Fig.~#1#3}
\Crefformat{figure}{#2Fig.~#1#3}
\crefformat{table}{#2Table~#1#3}
\Crefformat{table}{#2Table~#1#3}
\crefformat{section}{#2Sec.~#1#3}
\Crefformat{section}{#2Sec.~#1#3}
\crefformat{problem}{#2Problem~#1#3}
\Crefformat{problem}{#2Problem~#1#3}
\Crefmultiformat{figure}{Figs.~#2#1#3}{ and~#2#1#3}{, #2#1#3}{ and~#2#1#3}

\usepackage[caption=false]{subfig}

\usepackage{booktabs}
\title{\LARGE\bf Structured Pneumatic Fingerpads for Actively Tunable Grip Friction}

\author{Katherine Allison$^{a,b}$, Jonathan Kelly$^{a,\dagger}$, and Benjamin Hatton$^{b}$
\thanks{$^a$Authors are with the STARS Laboratory, University of Toronto Institute for Aerospace Studies, University of Toronto, Ontario, Canada. {\tt\footnotesize <firstname>.<lastname>@robotics.utias.utoronto.ca}}
\thanks{$^b$Authors are with the Bio-Inspired Materials \& Design Laboratory, Department of Materials Science \& Engineering, University of Toronto, Ontario, Canada.}
\thanks{$^\dagger$This research was supported in part by the Canada Research Chairs program.}}

\begin{document}
	
\maketitle 
\thispagestyle{empty}
\pagestyle{empty}

\begin{abstract} 
Grip surfaces with tunable friction can actively modify contact conditions, enabling transitions between higher- and lower-friction states for grasp adjustment.
Friction can be increased to grip securely and then decreased to gently release (e.g., for handovers) or manipulate in-hand. 
Recent friction-tuning surface designs using soft pneumatic chambers show good control over grip friction; however, most require complex fabrication processes and/or custom gripper hardware.
We present a practical structured fingerpad design for friction tuning that uses less than \$1 USD of materials, takes only seconds to repair, and is easily adapted to existing grippers.
Our design uses surface morphology changes to tune friction. 
The fingerpad is actuated by pressurizing its internal chambers, thereby deflecting its flexible grip surface out from or into these chambers.
We characterize the friction-tuning capabilities of our design by measuring the shear force required to pull an object from a gripper equipped with two independently actuated fingerpads.
Our results show that varying actuation pressure and timing changes the magnitude of friction forces on a gripped object by up to a factor of 2.8.
We demonstrate additional features including macro-scale interlocking behaviour and pressure-based object detection.
\end{abstract}

\section{Introduction}
\label{sec:introduction}

Successful manipulation relies on control over friction at the grip interface.
For most rigid grippers, stable grasps rely on force closure, in which contacts must be formed with sufficiently high friction forces to resist object wrenches \cite{PrattichizzoTrinkleGrasping2008}.
While friction can be increased using impactive grip forces (i.e., by squeezing harder), high impactive forces exerted by stiff surfaces may damage delicate objects \cite{HaoVisellSoftHandsEfficient2021,ShintakeEtAlSoftRoboticGrippers2018}. 
In contrast, soft grippers grasp more gently due to their high compliance, which both distributes impactive pressure and increases friction by increasing contact area \cite{ShintakeEtAlSoftRoboticGrippers2018,CutkoskyWrightFrictionStabilityDesign1986,CiocarlieEtAlSoftFingerModel2007,LuRojasSoftFingertips2019}. 

The challenge of securely grasping with reduced impactive force has motivated interdisciplinary work combining design principles from materials science and robotics.
Gripping surfaces can easily be modified at design time to increase their effective coefficient of friction (e.g., by adding surface texture \cite{HaoEtAlFrictionEnhancementFingerprintlike2024, MizushimaEtAlSurfaceTextureDeformable2017}).
However, increased friction is not always desirable---for example, high friction reduces the dexterity of soft grippers by inhibiting sliding \cite{LuEtAlOrigamiInspiredVariableFriction2020,RuotoloEtAlGraspingManipulationGeckoinspired2021,NojiriEtAlDevelopmentContactArea2019}. 
For versatile manipulation, the ability to actively tune friction is an attractive possibility. 
Grip surfaces with active friction tuning could use a higher-friction state to maintain a stable grasp and a lower-friction state when delicately placing an object or performing in-hand manipulation tasks.
Friction tuning has the potential to enable manipulation that is both controlled and precise, allowing for efficient handling and delicate adjustment.

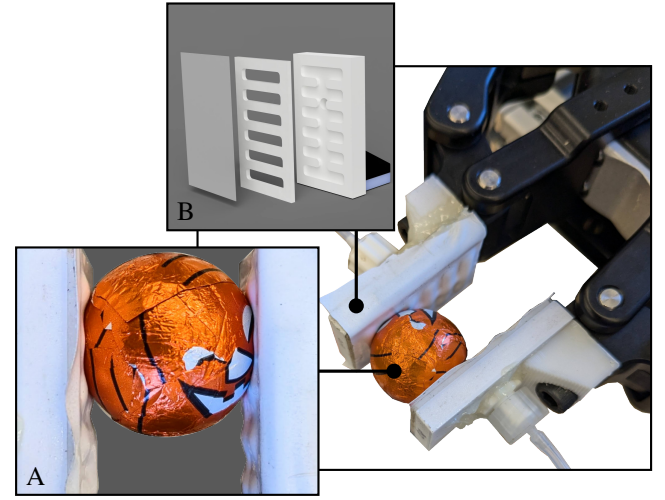
\begin{figure}[t]
    \centering
    \begin{tikzpicture}[]  
        \node (far) at (0,0){\includegraphics[width=6cm]{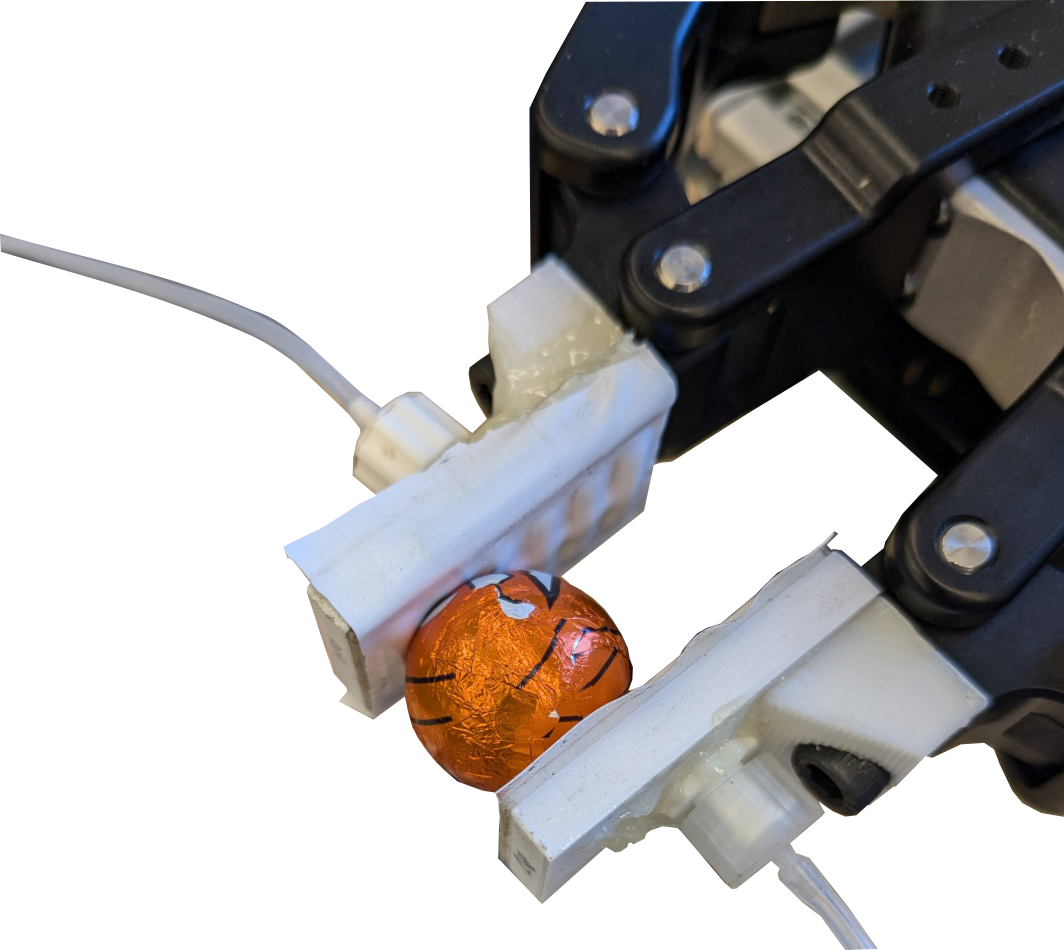}};
        \draw[thick] (-3,2.67) rectangle (3,-2.67);

        % inset 1: pic of ball
        \coordinate (p1) at (-5.4,-3);
        \coordinate (pc) at ($(p1)+{2/11}*(11,9)$);
        \coordinate (p2) at ($(p1)+{4/11}*(11,9)$);
        \draw[very thick] (pc) -- ($(pc)+(3,0)$);
        \fill ($(pc)+(3,0)$) circle (0.1);
        \fill[gray!70!black] (p1) rectangle (p2);
        \node (close) at (pc){\includegraphics[width=4cm]{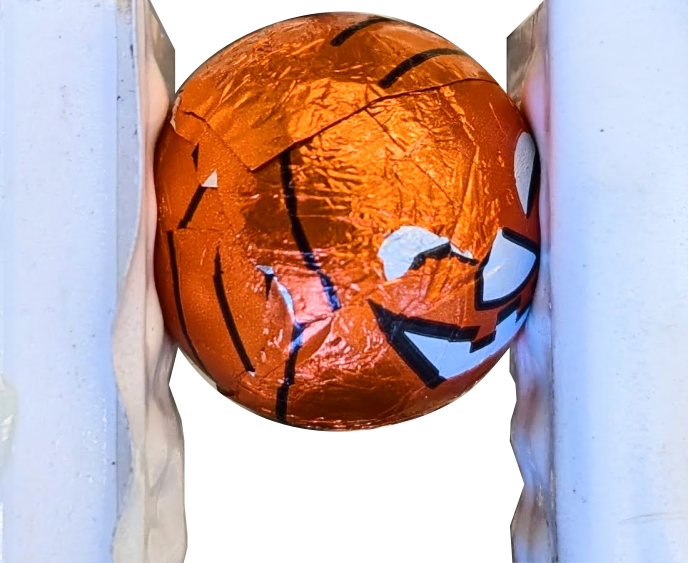}}; 
        \draw[thick] (p1) node[anchor=south west]{A} rectangle (p2);

        % inset 2: model
        \coordinate (q1) at ($(p1)+(2,3.5)$);
        \coordinate (qc) at ($(q1)+{1/2}*(3,3)$);
        \coordinate (q2) at ($(q1)+(3,3)$);
        \draw[very thick] ($(qc)+(1,0)$) -- ($(qc)+(1,-2.5)$);
        \fill ($(qc)+(1,-2.5)$) circle (0.1);
        \fill[white] (q1) rectangle (q2);
        \node (model) at (qc){\includegraphics[width=3cm]{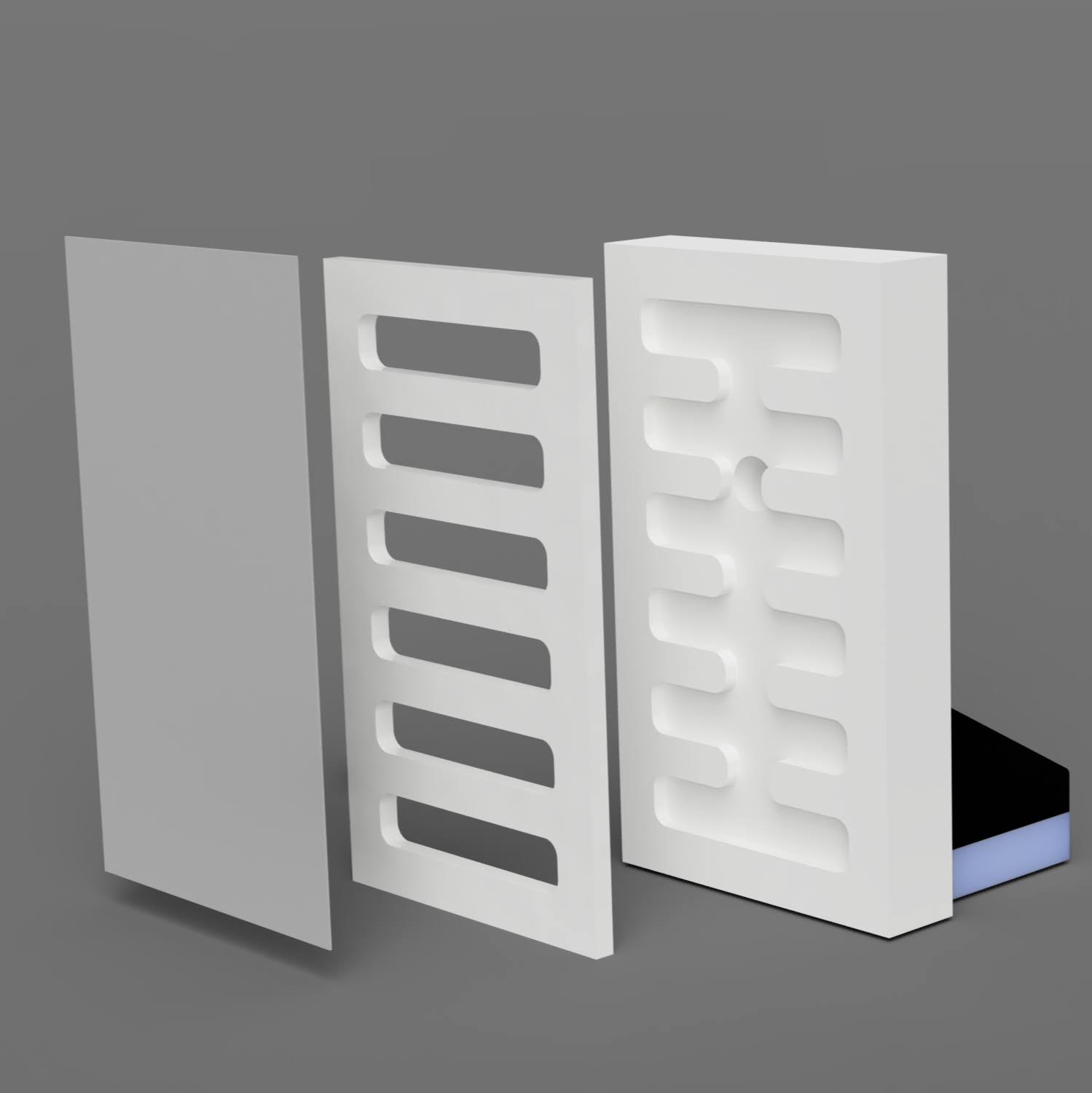}}; 
        \draw[thick] (q1) node[anchor=south west]{B} rectangle (q2);
    \end{tikzpicture}
    \caption{Structured pneumatic fingerpads gripping round object. Inset A: inflated active regions partially enveloping object. Inset B: render of fingerpad layers.} 
    \vspace{0.1mm}
    \label{fig:front}
    \vspace{-3mm}
\end{figure}

This work introduces an active friction tuning mechanism that is implemented with common prototyping materials and digital fabrication techniques.
Our design incorporates pneumatic channels to deflect structured active regions of a thin grip surface.
We embed these pneumatic channels directly into fingerpads that we attach to a commercially available rigid gripper.
By inflating or deflating the active regions of the grip surface, we can modify grip friction by modifying contact between the gripper and an object (e.g., by partially enveloping, as shown in \cref{fig:front}).

In this paper, we present the design and fabrication of these structured pneumatic fingerpads. 
We characterize the relationship between the pressure in the active regions and the friction forces on a gripped object. 
We show that actuating the fingerpads changes the magnitude of shear forces experienced by a gripped object during sliding.
Lastly, we illustrate how the fingerpads facilitate useful macro-scale interactions (e.g., interlocking) and demonstrate that their internal pressure enables a simple object detection scheme.

\section{Related Work}
\label{sec:related}

Pneumatic chambers have been incorporated in robotic gripper surfaces to achieve both macro-scale and micro-scale contact modification.
For instance, soft chambers have long been used to envelop or wrap around objects.
This macro-scale behaviour drastically increases contact area, sometimes even to the point of establishing form closure \cite{PrattichizzoTrinkleGrasping2008,ShintakeEtAlSoftRoboticGrippers2018}.
Although pneumatic chambers often make up entire grippers or gripper fingers (e.g., PneuNets \cite{IlievskiEtAlSoftRoboticsChemists2011}), the load and torque limitations faced by soft grippers have motivated the development of hybrid rigid-soft grippers as a compliant yet strong solution \cite{ChenEtAlSoftrigidCouplingGrippers2023}.

\begin{figure}[bt]
    \centering
    \begin{tikzpicture}
% INTERLOCKING
    \coordinate (cs) at (0,0);
    \coordinate (cw) at (0.7,0);
    \coordinate (cl) at (0,-1);
    \coordinate (cew) at (0.1,0);
    \coordinate (cel) at (0,-0.1);
    \coordinate (csx) at ($2*(cew)$);
    \coordinate (csy) at (cl);
    \coordinate (cbx) at ($3*(cew)$);
    \filldraw[fill=green!50!black] (cs) -- ($(cs)+(cel)$) -- ($(cs)+(cel)+(cew)$) -- ($(cs)+(csx)+(csy)$) -- ($(cs)+(cbx)+(cl)$) -- ($(cs)+(cw)-(cbx)+(cl)$) -- ($(cs)+(cw)-(csx)+(cl)$) -- ($(cs)+(cw)-(cew)+(cel)$) -- ($(cs)+(cw)+(cel)$) -- ($(cs)+(cw)$) -- cycle;

    \coordinate (fsl) at (-0.65,0.35);
    \coordinate (fsr) at (1.35,0.35);
    \coordinate (fwl) at (0.75,0);
    \coordinate (fwr) at (-0.75,0);
    \coordinate (fl) at (0,-1.7);
    \coordinate (arcstart) at (0,-0.2);
    \coordinate (arcsep) at (0,-0.5);
    \filldraw[fill=gray!5!white,draw=black] (fsl) -- ($(fsl)+(fwl)$) -- ($(fsl)+(fwl)+(arcstart)$) arc (90:270:0.15) -- ($(fsl)+(fwl)+(arcstart)+(arcsep)$)  arc (90:270:0.15) -- ($(fsl)+(fwl)+(arcstart)+2*(arcsep)$)  arc (90:270:0.15) -- ($(fsl)+(fwl)+(fl)$) -- ($(fsl)+(fl)$) -- cycle;
    \filldraw[fill=gray!5!white,draw=black] (fsr) -- ($(fsr)+(fwr)$) -- ($(fsr)+(fwr)+(arcstart)$) arc (90:-90:0.15) -- ($(fsr)+(fwr)+(arcstart)+(arcsep)$)  arc (90:-90:0.15) -- ($(fsr)+(fwr)+(arcstart)+2*(arcsep)$)  arc (90:-90:0.15) -- ($(fsr)+(fwr)+(fl)$) -- ($(fsr)+(fl)$) -- cycle;

    \coordinate (pic_offset) at (2.75,0);
    \coordinate (fsl2) at ($(fsl) + (pic_offset)$);
    \coordinate (fsr2) at ($(fsr) + (pic_offset)$);
    \filldraw[fill=gray!5!white,draw=black] (fsl2) -- ($(fsl2)+(fwl)$) -- ($(fsl2)+(fwl)+(arcstart)$) arc (90:-90:0.15) -- ($(fsl2)+(fwl)+(arcstart)+(arcsep)$)  arc (90:-90:0.15) -- ($(fsl2)+(fwl)+(arcstart)+2*(arcsep)$)  arc (90:-90:0.15) -- ($(fsl2)+(fwl)+(fl)$) -- ($(fsl2)+(fl)$) -- cycle;
    \filldraw[fill=gray!5!white,draw=black] (fsr2) -- ($(fsr2)+(fwr)$) -- ($(fsr2)+(fwr)+(arcstart)$) arc (90:270:0.15) -- ($(fsr2)+(fwr)+(arcstart)+(arcsep)$)  arc (90:270:0.15) -- ($(fsr2)+(fwr)+(arcstart)+2*(arcsep)$)  arc (90:270:0.15) -- ($(fsr2)+(fwr)+(fl)$) -- ($(fsr2)+(fl)$) -- cycle;
    
    \coordinate (c_shift) at (0.35,-0.25);
    \coordinate (cs2) at ($(cs) + (pic_offset) + (c_shift)$);
    \filldraw[fill=green!50!black] (cs2) circle (0.2);

    \coordinate (c_shiftx) at (0.11,0);
    \coordinate (fsl3) at ($(fsl) + 2*(pic_offset)$);
    \coordinate (fsr3) at ($(fsr) + 2*(pic_offset) + 2*(c_shiftx)$);
    \filldraw[fill=gray!5!white,draw=black] (fsl3) -- ($(fsl3)+(fwl)$) -- ($(fsl3)+(fwl)+(arcstart)$) arc (90:-90:0.15) -- ($(fsl3)+(fwl)+(arcstart)+(arcsep)$)  arc (90:-90:0.15) -- ($(fsl3)+(fwl)+(arcstart)+2*(arcsep)$)  arc (90:-90:0.15) -- ($(fsl3)+(fwl)+(fl)$) -- ($(fsl3)+(fl)$) -- cycle;
    \filldraw[fill=gray!5!white,draw=black] (fsr3) -- ($(fsr3)+(fwr)$) -- ($(fsr3)+(fwr)+(arcstart)$) arc (90:270:0.15) -- ($(fsr3)+(fwr)+(arcstart)+(arcsep)$)  arc (90:270:0.15) -- ($(fsr3)+(fwr)+(arcstart)+2*(arcsep)$)  arc (90:270:0.15) -- ($(fsr3)+(fwr)+(fl)$) -- ($(fsr3)+(fl)$) -- cycle;
    
    \coordinate (c_shifty) at (0,0.3);
    \coordinate (cs3) at ($(cs) + 2*(pic_offset) + (c_shiftx) + (c_shifty)$);
    \coordinate (cl3) at (0,-0.6);
    \filldraw[fill=green!50!black] (cs3) .. controls ($(cs3)+{1/2}*(cw)+{1/4}*(cl3)$) .. ($(cs3)+(cw)$) .. controls ($(cs3)+{0.7}*(cw)+{1/2}*(cl3)$) .. ($(cs3)+(cw)+(cl3)$)  .. controls ($(cs3)+{1/2}*(cw)+{3/4}*(cl3)$) .. ($(cs3)+(cl3)$)  .. controls ($(cs3)+{0.3}*(cw)+{1/2}*(cl3)$) .. cycle;

    \path ($(cs)+{1/2}*(cw)+(0,0.5)$) node {\footnotesize interlocking (convex)}
    ($(cs)+(pic_offset)+{1/2}*(cw)+(0,0.5)$) node {\footnotesize partial enveloping}
    ($(cs)+2*(pic_offset) +{1/2}*(cw)+(0,0.5)$) node {\footnotesize interlocking (concave)};        
\end{tikzpicture}
    \vspace{0.1mm}
    \caption{Different types of macro-scale contact modification.}
    \label{fig:macro}
    \vspace{-3mm}
\end{figure}
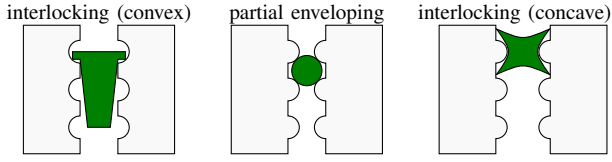

Some recent rigid-soft grippers augment rigid fingers with soft pneumatic fingerpads, which offer both passive compliance and active pressure-distributing and enveloping capabilities.
Most such designs use only a single chamber on each finger \cite{TrinhEtAlNovelRigidsoftGripper2024,KeelyEtAlCombiningDecouplingRigid2024,ChoiKocDesignFeasibilityTests2006} or phalange \cite{PozziEtAlSoftPneumaticPads2024}, in which the entire fingerpad surface forms a single active region (rather than creating multiple active regions as in \cite{HeEtAlSoftFingertipsTactile2020}). 
Although these large active regions do increase friction at the macro-scale by enveloping objects, their friction-tuning effects are coupled to the global morphology of the fingerpads. 
When the pads inflate, they form bubble shapes that can significantly laterally displace small objects near the sides of the pads.
Additionally, most designs apply only positive pneumatic pressure. 
The few that apply negative pressure (e.g., \cite{KeelyEtAlCombiningDecouplingRigid2024}) typically tune compliance or adhesion rather than friction. 
In contrast, our structured fingerpads use a single pneumatic input to actuate an array of small active regions. 
Actuation changes the local morphology of the active regions, tuning micro-scale friction with only minor changes to the global pad morphology and thus facilitating stable grasping of small objects. 
In certain cases, more macro-scale contact modification is possible: we can apply negative pressure to recess the active regions (enabling interlocking with convex features) or positive pressure to protrude the active regions (enabling partial enveloping of convex objects or interlocking with concave features) as depicted in \cref{fig:macro}. 

Drawing on investigation of tunable surfaces in materials science \cite{CalaisAlvaradoAdvancedFunctionalMaterials2019}, some recent grip surface designs use active pneumatic control to tune micro-scale contact properties.
For instance, Becker et al.\ use local morphology changes to tune friction in their soft gripper by inflating a high-friction elastomeric bladder until it protrudes through holes in a low-friction restraining layer \cite{BeckerEtAlTunableFrictionConstrained2017}.
This dual-layer design achieves an impressive increase in shear forces by a factor of five during object sliding. 
However, unlike our design, it is implemented in a fully soft gripper, cannot generally envelop or interlock with objects, and does not apply negative pressure. 
Tian et al.\ combine soft dry adhesives with a rigid base and use pneumatic control to deflect the adhesive surface in structured active regions \cite{TianEtAlGeckoEffectInspiredSoft2019}. 
This rigid-soft gripper increases adhesion force by a factor of 20; however, its dry adhesive is challenging to fabricate and requires specialized equipment. 
Moreover, the surface is designed to grip with adhesion alone and does not directly tune friction.
Trinh et al.\ design a theoretical rigid-soft fingerpad that, like our design, deflects a flexible surface in an array of small active regions \cite{TrinhEtAlTheoreticalFoundationDesign2019}. 
However, their design is not fabricated or tested experimentally and does not apply negative pressure. 
% For recent reviews of surfaces with adaptive contact properties in robotics, see \cite{ArztEtAlFunctionalSurfaceMicrostructures2021,XuEtAlLearningBiologicalAttachment2022,HassaniBajiRecentProgressUse2023,ZhaoEtAlReviewBioinspiredDry2024}, and especially \cite{CalaisAlvaradoAdvancedFunctionalMaterials2019}.
%
In contrast to most other tunable surfaces, our fingerpad design does not rely on moulded components. 
Rather, we leverage digitally fabricated and off-the-shelf components that minimize the hands-on fabrication workload and permit straightforward repair and customization. 
We also attach our fingerpads to a commercially available robotic gripper rather than using a custom gripper, demonstrating easy adaptation of our design to existing manipulators. 

\section{Pneumatic Fingerpad Design}
\label{sec:design}

\begin{figure}[t]
	\vspace{-\intextsep}
    \centering
    \subfloat[Exploded view of layers.]{\label{fig:design:CAD}
    \includegraphics[width=0.4\linewidth]{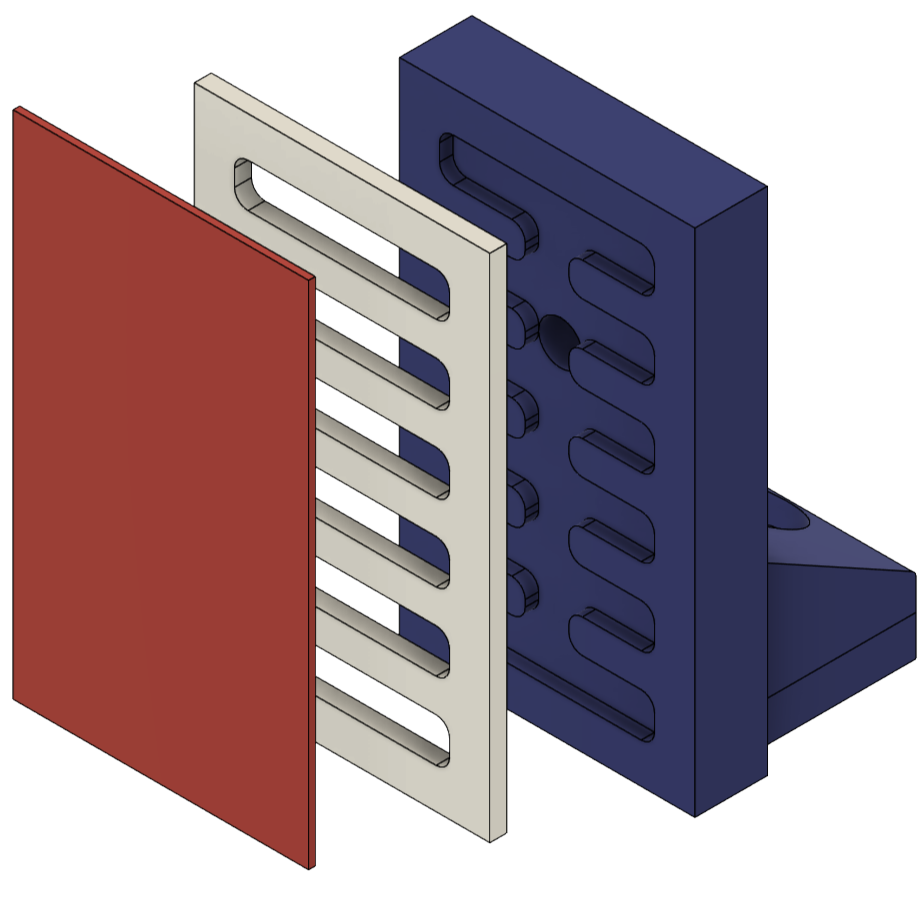}
    }%
    \subfloat[Diagram of fingerpad actuation states.]{\label{fig:design:schem}
    \begin{tikzpicture}[scale=2.25]         
    \coordinate (pic_offset) at (0,-0.6);
    
    % layer thicknesses           
    \coordinate (ft) at (0,-0.05);
    \coordinate (mt) at (0,-0.08);
    \coordinate (bt) at (0,-0.2);
    \coordinate (ct) at (${2/3}*(bt)$);

    % arc parameters
    \coordinate (arcstart) at (0.2,0);
    \coordinate (arcsep) at (0.5,0);
    \coordinate (arcd) at (0.282,0); % chord length

    % NEUTRAL
    \coordinate (fsn) at (0,0);
    \coordinate (fw) at (1.7,0); 
    \coordinate (fsnta) at ($(fsn) + (arcstart)$);
    \coordinate (fsnb) at ($(fsn) + (ft)$);
    \coordinate (fsnba) at ($(fsnb) + (arcstart)$); 
    
    \filldraw[fill=yellow!15!] (fsnb) -- (fsnba) -- ($(fsnba)+(mt)$) -- ($(fsnb)+(mt)$) -- cycle;
    \filldraw[fill=yellow!15!] ($(fsnb)+(arcsep)$) -- ($(fsnba)+(arcsep)$) -- ($(fsnba)+(mt)+(arcsep)$) -- ($(fsnb)+(mt)+(arcsep)$) -- cycle;
    \filldraw[fill=yellow!15!] ($(fsnb)+{2}*(arcsep)$) -- ($(fsnba)+{2}*(arcsep)$) -- ($(fsnba)+(mt)+{2}*(arcsep)$) -- ($(fsnb)+(mt)+{2}*(arcsep)$) -- cycle;
    \filldraw[fill=yellow!15!] ($(fsnb)+{3}*(arcsep)$) -- ($(fsnba)+{3}*(arcsep)$) -- ($(fsnba)+(mt)+{3}*(arcsep)$) -- ($(fsnb)+(mt)+{3}*(arcsep)$) -- cycle;

    \coordinate (fsnz) at ($(fsnb) + (mt)$);
    \coordinate (fsnza) at ($(fsnb) + (mt) + (arcstart)$);
    \filldraw[fill=base_blue] 
    (fsnz) -- ($(fsnza)$) -- ($(fsnza)+(ct)$) -- ($(fsnz)+(ct)+(arcsep)$) -- 
    ($(fsnz)+(arcsep)$) -- ($(fsnza)+(arcsep)$) -- ($(fsnza)+(arcsep)+(ct)$) -- ($(fsnz)+{2}*(arcsep)+(ct)$) -- 
    ($(fsnz)+{2}*(arcsep)$) -- ($(fsnza)+{2}*(arcsep)$) -- ($(fsnza)+{2}*(arcsep)+(ct)$) -- ($(fsnz)+{3}*(arcsep)+(ct)$) --
    ($(fsnz)+{3}*(arcsep)$) -- ($(fsnza)+{3}*(arcsep)$) -- ($(fsnza)+{3}*(arcsep)+(bt)$) -- 
    ($(fsnz)+(bt)$) -- cycle;
    
    \filldraw[fill=top_red,draw=black] (fsn)-- ($(fsn)+(fw)$) -- ($(fsn)+(ft)+(fw)$) -- ($(fsn)+(ft)$)  -- cycle;

    \path 
    ($(fsn)$) node[anchor=south east] {$P_0$}
    ($(fsnza)$) node[anchor=west] {$\scriptstyle P_N$}
    ($(fsnza)+(arcsep)$) node[anchor=west] {$\scriptstyle P_N$}
    ($(fsnza)+{2}*(arcsep)$) node[anchor=west] {$\scriptstyle P_N$}
    ($(fsn)+{1/2}*(fw)$) node[anchor=south] {neutral state ($P_N=P_0$)};

    % DEFLATED
    \coordinate (fsdt) at ($(fsn) + (pic_offset)$);
    \coordinate (fsdta) at ($(fsdt) + (arcstart)$);
    \coordinate (fsdb) at ($(fsdt) + (ft)$);
    \coordinate (fsdba) at ($(fsdb) + (arcstart)$);
    
    \filldraw[fill=yellow!15!] (fsdb) -- (fsdba) -- ($(fsdba)+(mt)$) -- ($(fsdb)+(mt)$) -- cycle;
    \filldraw[fill=yellow!15!] ($(fsdb)+(arcsep)$) -- ($(fsdba)+(arcsep)$) -- ($(fsdba)+(mt)+(arcsep)$) -- ($(fsdb)+(mt)+(arcsep)$) -- cycle;
    \filldraw[fill=yellow!15!] ($(fsdb)+{2}*(arcsep)$) -- ($(fsdba)+{2}*(arcsep)$) -- ($(fsdba)+(mt)+{2}*(arcsep)$) -- ($(fsdb)+(mt)+{2}*(arcsep)$) -- cycle;
    \filldraw[fill=yellow!15!] ($(fsdb)+{3}*(arcsep)$) -- ($(fsdba)+{3}*(arcsep)$) -- ($(fsdba)+(mt)+{3}*(arcsep)$) -- ($(fsdb)+(mt)+{3}*(arcsep)$) -- cycle;

    \coordinate (fsdz) at ($(fsdb) + (mt)$);
    \coordinate (fsdza) at ($(fsdb) + (mt) + (arcstart)$);
    \filldraw[fill=base_blue] 
    (fsdz) -- ($(fsdza)$) -- ($(fsdza)+(ct)$) -- ($(fsdz)+(ct)+(arcsep)$) -- 
    ($(fsdz)+(arcsep)$) -- ($(fsdza)+(arcsep)$) -- ($(fsdza)+(arcsep)+(ct)$) -- ($(fsdz)+{2}*(arcsep)+(ct)$) -- 
    ($(fsdz)+{2}*(arcsep)$) -- ($(fsdza)+{2}*(arcsep)$) -- ($(fsdza)+{2}*(arcsep)+(ct)$) -- ($(fsdz)+{3}*(arcsep)+(ct)$) --
    ($(fsdz)+{3}*(arcsep)$) -- ($(fsdza)+{3}*(arcsep)$) -- ($(fsdza)+{3}*(arcsep)+(bt)$) -- 
    ($(fsdz)+(bt)$) -- cycle;
    
    \filldraw[fill=top_red,draw=black] (fsdt)-- ($(fsdta)$) arc (-160:-20:0.15) -- ($(fsdta)+(arcsep)$)  arc (-160:-20:0.15) -- ($(fsdta)+2*(arcsep)$)  arc (-160:-20:0.15) -- ($(fsdt)+(fw)$) -- ($(fsdb)+(fw)$) -- ($(fsdba)+2*(arcsep)+(arcd)$)  arc (-20:-160:0.15) -- ($(fsdba)+(arcsep)+(arcd)$)  arc (-20:-160:0.15) -- ($(fsdba)+(arcd)$)  arc (-20:-160:0.15) -- ($(fsdb)$)  -- cycle;

    \path 
    ($(fsdza)+(0,-0.07)$) node[anchor=west] {$\scriptstyle P_D$}
    ($(fsdza)+(arcsep)+(0,-0.07)$) node[anchor=west] {$\scriptstyle P_D$}
    ($(fsdza)+{2}*(arcsep)+(0,-0.07)$) node[anchor=west] {$\scriptstyle P_D$}
    ($(fsdt)+{1/2}*(fw)$) node[anchor=south] {deflated state ($P_D<P_0$)};

    % INFLATED
    \coordinate (fsit) at ($(fsn) + 2*(pic_offset)-(0,0.08)$);
    \coordinate (fsita) at ($(fsit) + (arcstart)$);
    \coordinate (fsib) at ($(fsit) + (ft)$);
    \coordinate (fsiba) at ($(fsib) + (arcstart)$);
    
    \filldraw[fill=yellow!15!] (fsib) -- (fsiba) -- ($(fsiba)+(mt)$) -- ($(fsib)+(mt)$) -- cycle;
    \filldraw[fill=yellow!15!] ($(fsib)+(arcsep)$) -- ($(fsiba)+(arcsep)$) -- ($(fsiba)+(mt)+(arcsep)$) -- ($(fsib)+(mt)+(arcsep)$) -- cycle;
    \filldraw[fill=yellow!15!] ($(fsib)+{2}*(arcsep)$) -- ($(fsiba)+{2}*(arcsep)$) -- ($(fsiba)+(mt)+{2}*(arcsep)$) -- ($(fsib)+(mt)+{2}*(arcsep)$) -- cycle;
    \filldraw[fill=yellow!15!] ($(fsib)+{3}*(arcsep)$) -- ($(fsiba)+{3}*(arcsep)$) -- ($(fsiba)+(mt)+{3}*(arcsep)$) -- ($(fsib)+(mt)+{3}*(arcsep)$) -- cycle;

    \coordinate (fsiz) at ($(fsib) + (mt)$);
    \coordinate (fsiza) at ($(fsib) + (mt) + (arcstart)$);
    \filldraw[fill=base_blue] 
    (fsiz) -- ($(fsiza)$) -- ($(fsiza)+(ct)$) -- ($(fsiz)+(ct)+(arcsep)$) -- 
    ($(fsiz)+(arcsep)$) -- ($(fsiza)+(arcsep)$) -- ($(fsiza)+(arcsep)+(ct)$) -- ($(fsiz)+{2}*(arcsep)+(ct)$) -- 
    ($(fsiz)+{2}*(arcsep)$) -- ($(fsiza)+{2}*(arcsep)$) -- ($(fsiza)+{2}*(arcsep)+(ct)$) -- ($(fsiz)+{3}*(arcsep)+(ct)$) --
    ($(fsiz)+{3}*(arcsep)$) -- ($(fsiza)+{3}*(arcsep)$) -- ($(fsiza)+{3}*(arcsep)+(bt)$) -- 
    ($(fsiz)+(bt)$) -- cycle;
    
    \filldraw[fill=top_red,draw=black] (fsit)-- ($(fsita)$) arc (160:20:0.15) -- ($(fsita)+(arcsep)$)  arc (160:20:0.15) -- ($(fsita)+2*(arcsep)$)  arc (160:20:0.15) -- ($(fsit)+(fw)$) -- ($(fsib)+(fw)$) -- ($(fsiba)+2*(arcsep)+(arcd)$)  arc (20:160:0.15) -- ($(fsiba)+(arcsep)+(arcd)$)  arc (20:160:0.15) -- ($(fsiba)+(arcd)$)  arc (20:160:0.15) -- ($(fsib)$)  -- cycle;

    \path 
    ($(fsiza)$) node[anchor=west] {$\scriptstyle P_I$}
    ($(fsiza)+(arcsep)$) node[anchor=west] {$\scriptstyle P_I$}
    ($(fsiza)+{2}*(arcsep)$) node[anchor=west] {$\scriptstyle P_I$}
    ($(fsit)+{1/2}*(fw)+(0,0.08)$) node[anchor=south] {inflated state ($P_I>P_0$)};       
\end{tikzpicture} 
    }%
    \caption{Exploded model of fingerpad layers (with red top layer, white middle layer, and blue base layer) with schematic showing the behaviour of the top layer in each actuation state.}
    \label{fig:design}
    \vspace{-3mm}
\end{figure}
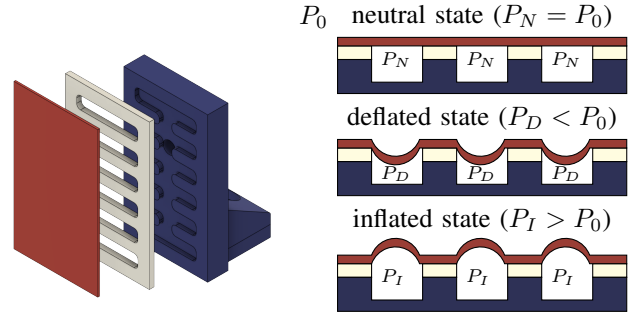

We propose a structured soft fingerpad design that uses pneumatic actuation to locally deflect the grip surface, thereby tuning friction at the grip interface. Our fingerpads are constructed by stacking three flat layers:
\begin{enumerate}
    \item A rigid base layer with an engraved channel that connects to a pneumatic input via an inlet fitting.
    \item An adhesive middle layer with cutouts that define the active region geometry.
    \item A thin, flexible top layer that acts as the grip contact surface and is deflected in the active regions by pneumatic pressure in the chamber below.
\end{enumerate}
An exploded model of these layers is pictured in \cref{fig:design:CAD}.

\subsection{Principle of Operation}

\begin{figure}[t!]
    \vspace{-\intextsep + 0.7mm}
    \centering
    \subfloat[Neutral at 0 kPa.]{\label{fig:states:neutral}
    \includegraphics[width=0.3\linewidth]{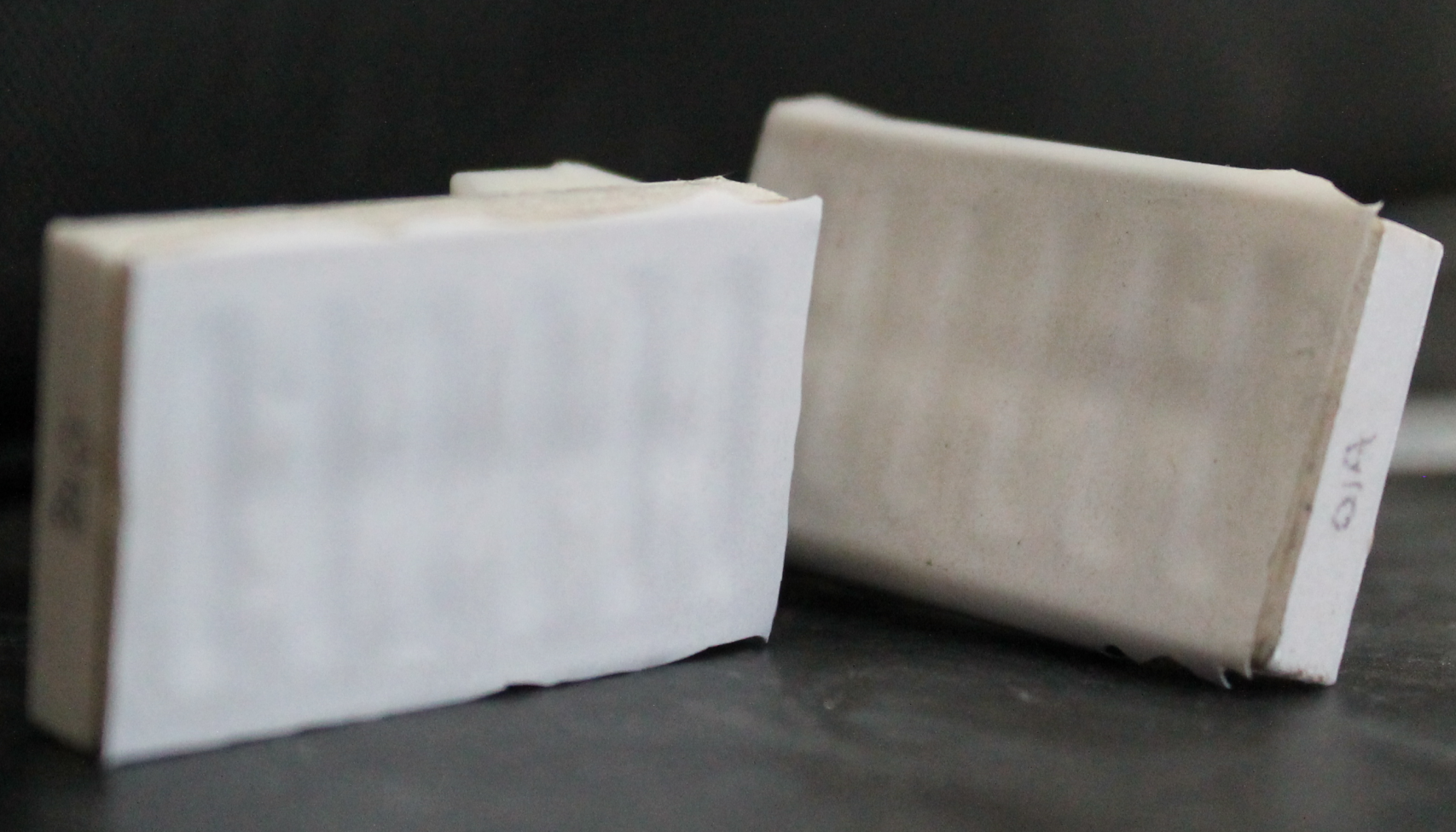}
    }%
    \subfloat[Deflated at -40 kPa.]{\label{fig:states:deflated}
    \includegraphics[width=0.3\linewidth]{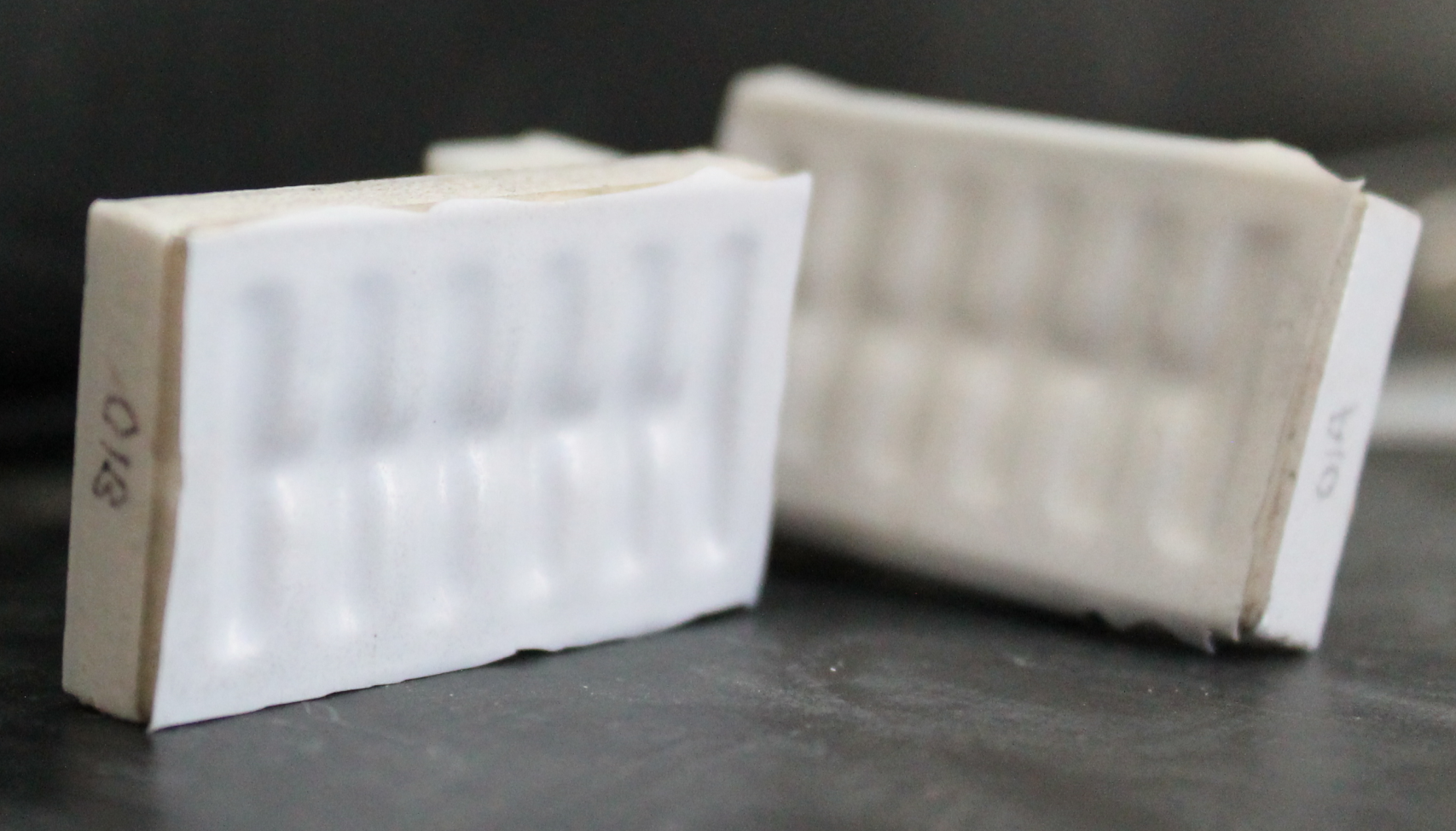}
    }%
    \subfloat[Inflated at 40 kPa.]{\label{fig:states:inflated}
    \includegraphics[width=0.3\linewidth]{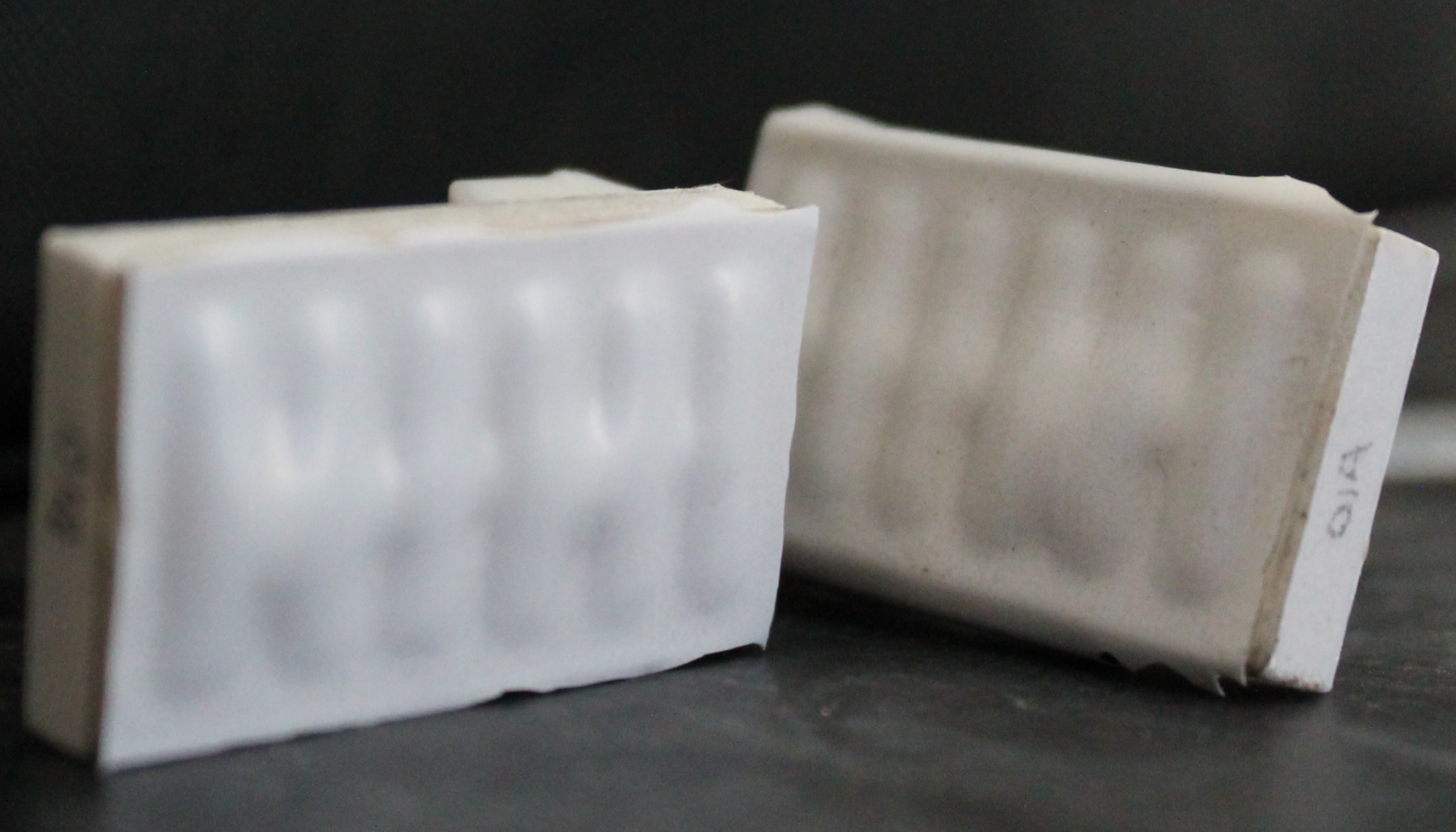}
    }%
    \caption{Pictures of fingerpads in different actuation states. These fingerpads were used in over 250 experimental test cycles and show signs of wear as delamination between some active regions.}
    \label{fig:states}
    \vspace{-3mm}
\end{figure}

The three fingerpad layers form a pneumatic chamber with three actuation states (shown schematically in \cref{fig:design:schem}):
\begin{enumerate}
    \item \textit{Neutral}: internal pressure $P_N$ is equal to atmospheric pressure $P_0$, so active regions of the top layer are coplanar with inactive regions (as in \cref{fig:states:neutral}). 
    \item \textit{Deflated}: internal pressure $P_D$ is less than $P_0$, so active regions recess into the chamber (as in \cref{fig:states:deflated}). 
    \item \textit{Inflated}: internal pressure $P_I$ is greater than $P_0$, so active regions protrude from the chamber (as in \cref{fig:states:inflated}).
\end{enumerate}
The actual pressure levels applied vary depending on the elastic modulus of the top layer, the geometry of the active regions, and the desired changes in frictional properties.

\subsection{Design and Fabrication Process}

\begin{table}[b!]
    \caption{Design and fabrication parameters}
    \label{tab:design}
    \centering
    \renewcommand{\arraystretch}{1.2}
    \begin{tabular}{ll}
    \toprule
    \textbf{Parameter} & \textbf{Design Specification}\\
    \midrule
    Total Length & 38 mm\\
    Total Width & 22 mm\\
    Total Thickness & 6.9 mm\\
    Active Region Length & 16 mm\\
    Active Region Width & 3.25 mm\\
    \midrule
    Base Layer Material & Extruded Acrylic \\
    Base Layer Material Cost & USD 0.07 \\
    \midrule
    Middle Layer Material & 3M\texttrademark{} VHB\texttrademark{} Tape\\
    Middle Layer Material Cost & USD 0.33 \\
    \midrule
    Surface Layer Material & PVC Tape \\
    Surface Layer Material Cost & USD 0.01 \\
    Surface Layer Thickness & 0.16 mm\\
    \bottomrule
    \end{tabular}
\end{table}

The fingerpad design and fabrication parameters are summarized in \cref{tab:design}.
The primary design challenge is the selection of the material to use for the top layer.
This material must be flexible enough to deflect under pneumatic pressure across a small area and strong enough to resist tearing under shear load. 
It must also have enough surface energy to adhere firmly to the middle layer at the active region boundaries.
We chose to use commercially available tapes to simplify fabrication, lower costs, and improve the inter-layer adhesive connection.
After testing different types of tape, we selected polyvinyl chloride (PVC) tape based on its excellent elongation and high strength.

The middle layer both holds the fingerpad assembly together and defines its active regions. 
We selected a double-sided acrylic foam tape based on its high strength and compatibility with laser cutting.
To specify the active region geometry, we used a trial-and-error approach to determine (1) a width that allowed the top layer to visibly deflect and (2) a separation distance that ensured good inter-layer adhesion and prevented delamination.
We then designed the active region layout and laser cut middle layer pieces in batches. 

The channel in the base layer controls airflow from a pneumatic input to the active regions. 
We engraved channels into acrylic pieces using a desktop CNC milling machine. 
The runtime was less than 5 minutes per piece.
We drilled a through-hole in each piece and inserted a Luer lock fitting, which we then secured and sealed with two-part epoxy.
Finally, we 3D-printed small connectors (for use in mounting the fingerpads to gripper fingers). 
We secured the connectors to the backs of the base layer pieces, again using epoxy.

\begin{figure}[t!]
    \centering
    \includegraphics[width=0.7\linewidth]{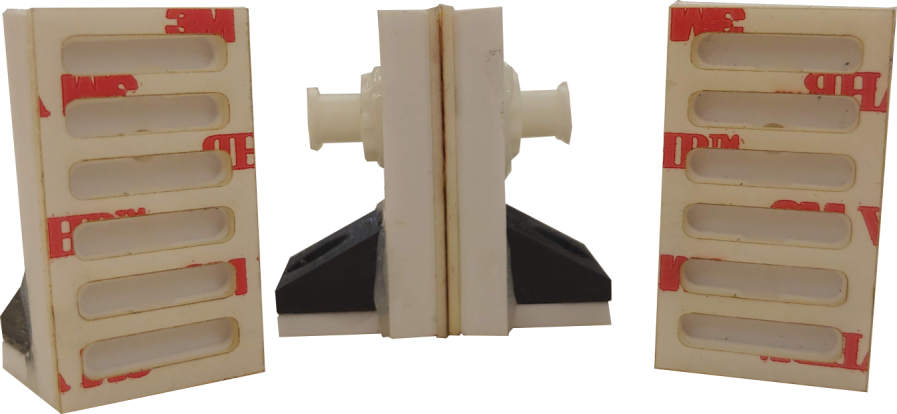}
    \vspace{-0.1mm}
    \caption{Several partially assembled fingerpads.}
    \label{fig:fab_partial}
    \vspace{-3mm}
\end{figure}

To assemble each structured pneumatic fingerpad, we first pressed the adhesive middle layer against the base layer (as in \cref{fig:fab_partial}), then pressed a strip of PVC tape on top of the middle layer. 

The rigid base layers of the fingerpads have been used for up to 250 test cycles and have not yet failed. 
After around 200 test cycles, the soft top and middle layers of the fingerpads started to show signs of wear (\Cref{fig:states}). 
However, repair is simple and fast---the top two layers can be peeled off and replaced in a matter of seconds. 
We can also easily adjust the active region geometry by redesigning and replacing the middle layer. 
Provided that the lengthscale of the new active regions is not significantly changed, we can even retain the same base layer.
Similarly, our fabrication process can be readily adapted to different equipment or material choices. 
For instance, the middle layers could be cut by hand, or the base layers could be 3D printed rather than milled. 

\section{Experimental Methodology}
\label{sec:methodology}

We used two plywood blocks as target objects in our experiments. 
We chose to use blocks because their relatively flat and parallel sides kept the contact area between the fingerpads and the target surface fairly consistent during testing. 
One block was wrapped with thin PTFE (Teflon) tape and served as a low-friction target, while the other was left unwrapped and served as a high-friction target. 

\begin{figure}[t!]
    \centering
    \subfloat[A FG-3008 force gauge travels on an actuated linear axis, exerting shear force on the gripped block via a connecting wire.]{\label{fig:setup:pic}
     \adjustbox{
    	rotate=5,
    	Clip=50mm 134mm 50mm 154mm,
 		width=0.9\linewidth}{
 		\includegraphics{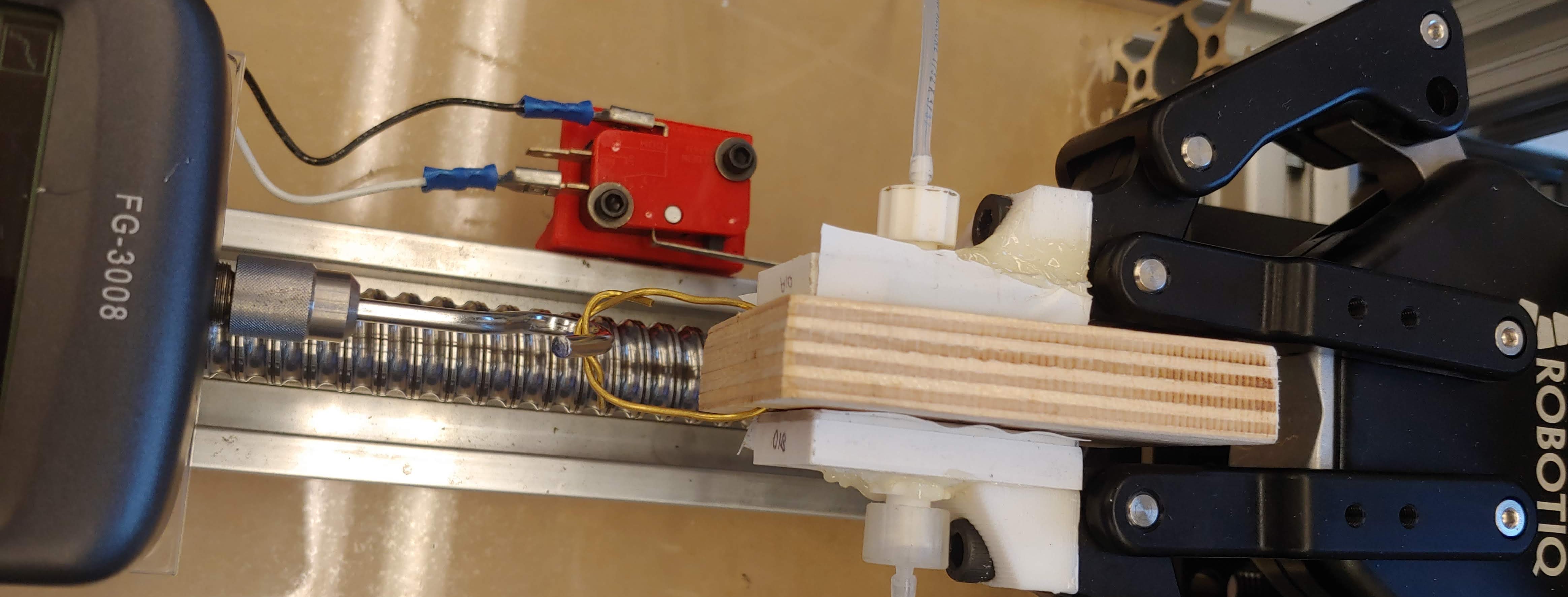}}
    }%
    \vspace{3mm}
    \subfloat[Schematic of gauge, block, and gripper during shear testing.]{\label{fig:setup:schematic}
    \begin{tikzpicture}[scale=0.9]
    \filldraw[fill=blue!10!gray!20!white] (0,0) rectangle (2.5,2) node[midway,align=center] {FG-3008\\gauge};
    \filldraw[fill=ply] (3,0.5) rectangle (4.75,1.5) node[midway] {block};
    \filldraw[fill=ply!5!white] (3.5,0) rectangle (4.5,0.5) node[midway] {pad};
    \filldraw[fill=ply!5!white] (3.5,1.5) rectangle (4.5,2) node[midway] {pad};
    \filldraw[fill=gray!50!black] (4.5,2) -- (4.5,1.5) -- (5,1.5)  to[out=0, in=0]  node[white,align=center,xshift=15]{2F-85\\gripper} (5,0.5) -- (4.5,0.5) -- (4.5,0) -- (6,0) to[out=0, in=0] (6,2) -- cycle;

    \draw[->, thick] (0,1) -- node[align=center]{travel\\direction}(-1.5,1);
    \draw[very thick] (2.5,1) -- (3,1);
\end{tikzpicture}
    }%
    \caption{Picture and schematic of shear testing apparatus.}
    \label{fig:setup}
    \vspace{-3mm}
\end{figure}
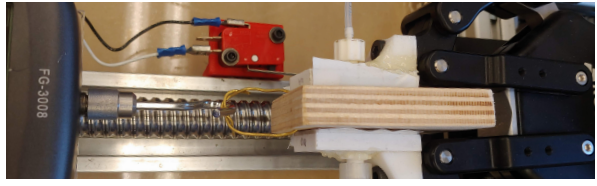
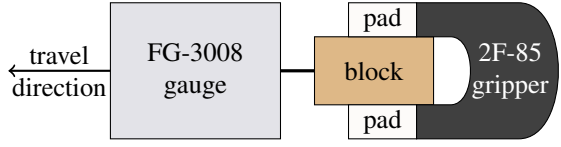

To assess the friction-tuning capabilities of the fingerpads, we used a custom-built force tester (pictured in \cref{fig:setup}) to measure the shear force required to cause a block to slip when gripped using different fingerpad actuation test cases. 
We used a custom-built pneumatic controller to actuate the fingerpads either before or after gripping (depending on the specific test case). 

We designed our fingerpads to replace the stock fingerpads of a 2F-85 adaptive gripper, an underactuated 2-fingered rigid gripper from Robotiq \cite{Robotiq2F852F1402018}. 
When gripping, the 2F-85 closes until \textit{either} a target finger separation distance is reached \textit{or} an object is detected (based on finger force feedback). 
We used a light grip during testing in order to fairly represent situations in which delicate objects must be grasped without excessive impactive force. 
We achieved this by setting the target finger separation distance to be slightly larger than the block width, thereby avoiding excessive grip forces caused by slight positional errors. 
Using this separation distance for every test gave us a consistently loose but stable grasp.
 
To set up each test, we mounted the gripper in a fixed position, closed its fingers to grip a block, and then connected the gripped block to a force gauge. 
We then moved the gauge and block away from the gripper at a constant velocity of 10 mm/s using a single-axis linear stage while taking frequent force measurements. 

\section{Experimental Results and Discussion}
\label{sec:experiments}

When conducting experiments, we used five fingerpad actuation test cases with differing pressure levels and timing:
\begin{enumerate}
    \item The neutral case (N), where the pads were not actuated.
    \item The deflated-live case (DL), where the pads were deflated \textit{after} gripping a block in the neutral state.
    \item The deflated-prior case (DP), where the pads were deflated \textit{before} gripping a block.
    \item The inflated-live case (IL), where the pads were inflated \textit{after} gripping a block in the neutral state.
    \item The inflated-prior case (IP), where the pads were inflated \textit{before} gripping a block.
\end{enumerate}

\subsection{Characterizing Force-Pressure Relationship}

\begin{figure}
    \centering
    \includegraphics[width=0.9\linewidth,trim=0 6pt 0 0,clip]{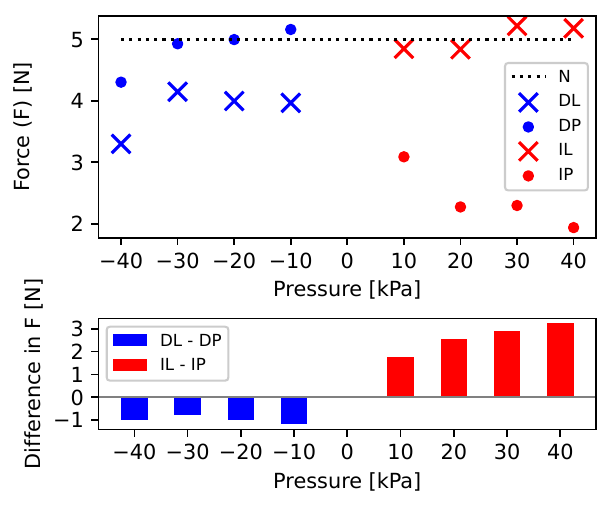}
    \caption{Shear force on a PTFE-wrapped block when gripped with fingerpads at different pressure levels. Forces in the IL case are much higher than in the IP case, and the difference between the cases (IL-IP) increases with pressure. In contrast, the difference between the DL and DP cases (DL-DP) is smaller and does not notably change with pressure. Each point is the average of data from 3 trials.}
    \label{fig:results:forcevspressure}
    \vspace{-3mm}
\end{figure}

We characterized the relationship between the internal pressure of the fingerpads and the resulting shear force on a gripped PTFE-wrapped block. 
To do so, we conducted experiments using pressure levels ranging from -40 kPa to +40 kPa. 
For each nonzero internal pressure level, we tested both the live case (DL and IL) and the prior case (DP and IP). 
The characterization results are shown in \cref{fig:results:forcevspressure}. 
The characterization testing exposed two general trends.

For the deflated states, the average shear forces were almost always lower in the DL case than in the DP case, with differences in forces (DL-DP) ranging from -0.2 N to 1.2 N. 
This means that the friction on the block is lower when the active regions deflate after gripping (as opposed to before). 
As pressure increased from -40 kPa to 0 kPa, there was a slight trend toward higher forces (for both DL and DP), likely as a result of increasing contact area (cA) between the fingerpad surface and the gripped block.

For the inflated states, on the other hand, the average shear forces in the IL case were always higher than in the IP case, with differences (IL-IP) ranging from 0.5 N to 3.2 N. 
This means that the friction on the block is higher when the active regions inflate after gripping (as opposed to before). 
As pressure increased from 0 kPa to +40 kPa, there were significant trends toward higher forces in the IL case and lower forces in the IP case. 
Again, these trends are likely related to changes in cA. 
In the IP case, the gripper detected an object soon after the protruding active regions of the fingerpads contacted the block, causing the gripper to stop closing before the inactive regions of the fingerpads came into contact. 
This resulted in a decrease in final grip cA relative to the neutral case.
In contrast, in the IL case, the inflation of the active regions likely both increased impactive pressure and increased cA by filling in any remaining gaps between the block and the active regions. 

\subsection{Testing Shear Force for Different Target Materials}

\begin{figure}
    \centering
    \includegraphics[width=0.9\linewidth,trim=0 6pt 0 0,clip]{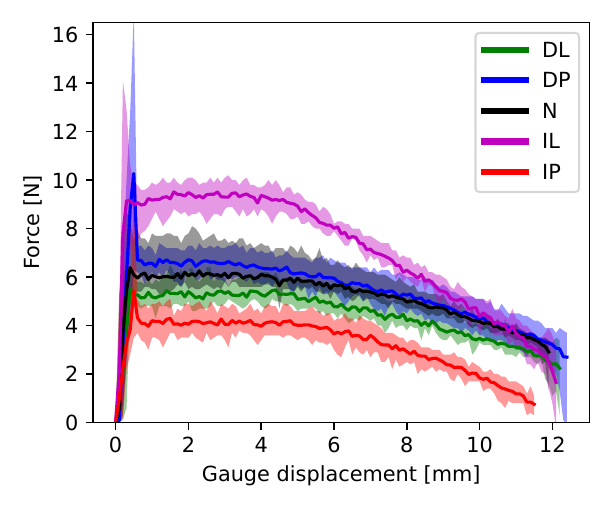}
    \vspace{-2mm}
    \caption{Shear force on a bare plywood block when gripped using different actuation cases. For each case, the shaded region is the area covered by force curves from 6 different trials and the solid line is the average of these 6 curves (computed as the average force at each gauge displacement value). Forces in the IL case are significantly higher than other cases. Forces in the IP case are significantly lower than other cases. Other cases (DL, DP, N) are generally similar to one another.}
    \label{fig:results:typical}
    \vspace{-3mm}
\end{figure}

We conducted more extensive testing of the shear force on gripped objects using a smaller set of actuation pressure levels: -40 kPa (DL or DP), 0 kPa (N), and 40 kPa (IL or IP). 
For these tests, we used both target blocks (bare plywood and PTFE-covered plywood). 
The mean maximum (peak) and steady-state (average) shear force values are reported for each target and actuation case in \cref{tab:forces}.
Data from a typical set of shear force tests are shown in \cref{fig:results:typical}. 

\begin{table}[b]
    \caption{Average ($F_a$) and peak ($F_p$) shear force measurements by target block material and actuation case$^{\mathrm{*}}$}
    \label{tab:forces}
    \centering
    \begin{tabular}{cccrrr}
    \toprule
    \textbf{Target} & \textbf{Case} & \textbf{Pressure} [kPa] & $F_a$ [N] & $F_p$ [N] & $F_a/F_0^{\mathrm{**}}$\\
    \midrule
    Plywood & DP & -40 & 6.6 & 9.7 &1.1\\
    Plywood & DL & -40 & 5.3 & 6.1 & 0.9\\
    Plywood & N & 0 & $F_0$=6.1 & 7.3 & 1\\
    Plywood & IP & +40 & 4.0 & 5.5 & 0.7\\
    Plywood & IL & +40 & 9.3 & 11.3 & 1.5\\
    \midrule
    PTFE & DP & -40 & 4.2 & 6.1 & 1\\
    PTFE & DL & -40 & 3.8& 5.0 & 0.9\\
    PTFE & N & 0 & $F_0$=4.4& 5.2 & 1\\
    PTFE & IP & +40 & 2.0& 2.7 & 0.5\\
    PTFE & IL & +40 & 5.6& 8.1 & 1.3\\
    \bottomrule\\
    \multicolumn{6}{l}{$^{\mathrm{*}}$ Each value is the average of 6 trials. For each trial, the peak force }\\
    \multicolumn{6}{l}{was computed using all data and the average was computed using data }\\
    \multicolumn{6}{l}{from a two-second interval in which the block was sliding steadily.}\\
    \multicolumn{6}{l}{$^{\mathrm{**}}$ Ratio of average force $F_a$ to $F_0$ (the average force measured for }\\
    \multicolumn{6}{l}{the same target block with fingerpads in neutral state).}
    \end{tabular}
    \vspace{-2mm}
\end{table}

Among the five actuation cases, we found that only the cases involving an inflated state (IL, IP) noticeably impacted shear friction forces compared to the neutral (N) case. 
By contrast, the cases involving a deflated state (DL, DP) did not substantially differ from the neutral (N) case. 

In the IL case, a gripped block experienced higher shear forces. 
In the IP case, a gripped block experienced lower shear forces. 
These trends match the characterization results: higher cA and impactive pressure in the IL case yields higher friction whereas lower cA in the IP case yields lower friction. 
The difference in friction forces between the IL and IP cases is substantial, with average force increased by a factor of 2.3 on bare plywood and by a factor of 2.8 on PTFE. 

The difference in friction forces between the IL and N cases is also notable. 
For bare plywood, the average force in the IL case was increased by a factor of 1.5 compared to the N case. 
For PTFE, the average force in the IL case was increased by a factor of 1.3 compared to the N case. 
Overall, it is clear that inflating these structured pneumatic fingerpads substantially impacts friction at the grip interface.

\subsection{Demonstration: Interlocking and Enveloping}

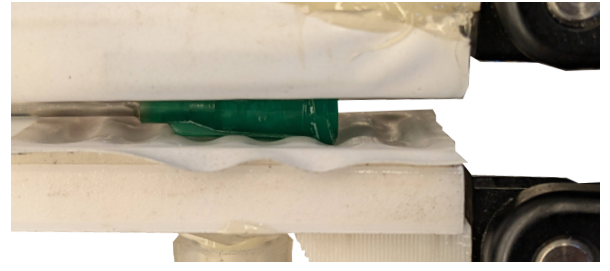
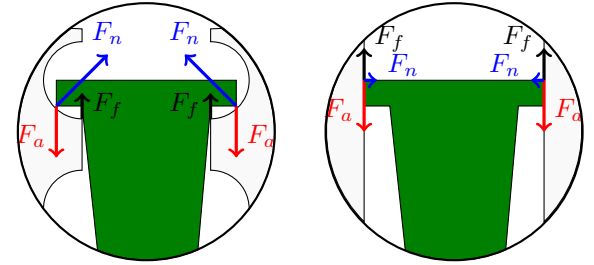
\begin{figure}[t!]
    \centering
    \subfloat[Interlocking between object and deflated fingerpads as seen from above.]{\label{fig:interlock:above}
    \adjustbox{
    	rotate=2,
    	Clip=2.5mm 2.3mm 2.5mm 7.1mm,
 		width=0.9\linewidth}{
    \includegraphics[width=0.7\linewidth]{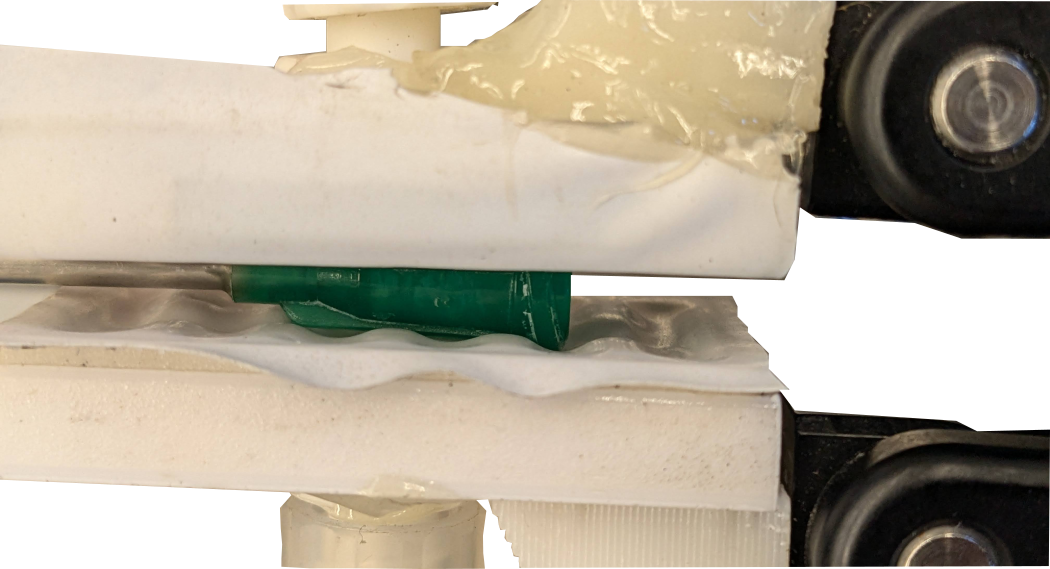}}
    }%
    \vspace{-1mm}
    \subfloat[Force diagram, interlocking.]{\label{fig:interlock:force}
    \begin{tikzpicture}[scale=1.7]
    \coordinate (cs_ins) at (0,0);
    \coordinate (cw) at (1.4,0);
    \coordinate (cl) at (0,-2);
    \coordinate (cew) at (0.2,0);
    \coordinate (cel) at (0,-0.2);
    \coordinate (csx) at ($2*(cew)$);
    \coordinate (csy) at (cl);
    \filldraw[fill=green!50!black] (cs_ins) -- ($(cs_ins)+(cel)$) -- ($(cs_ins)+(cel)+(cew)$) -- ($(cs_ins)+{0.8}*(csx)+{0.66}*(csy)$) -- ($(cs_ins)+{0.8}*(csx)+{0.66}*(csy)$) arc (247:293:1) -- ($(cs_ins)+(cw)-(cew)+(cel)$) -- ($(cs_ins)+(cw)+(cel)$) -- ($(cs_ins)+(cw)$) -- cycle;

    \coordinate (ins_start) at (0.2,0.4);
    \coordinate (arcstart_ins) at (0,-0.1);
    \coordinate (f_sep) at (1,0);
    \coordinate (fsl_ins) at ($(cs_ins)+(ins_start)$);
    \coordinate (fsr_ins) at ($(cs_ins)+(ins_start)+(f_sep)$);
    \coordinate (arcsep) at (0,-1);
    \coordinate (arcendl) at (-0.3,-0.3);
    \coordinate (arcendr) at (0.3,-0.3);
    \filldraw[fill=gray!5!white,draw=black] (fsl_ins) --  ($(fsl_ins)+(arcstart_ins)$) arc (90:270:0.3) -- ($(fsl_ins)+(arcstart_ins)+(arcsep)$)  arc (90:180:0.3) -- ($(fsl_ins)+(arcstart_ins)+(arcsep)+(arcendl)$) arc (216.5:126:1) -- cycle;
    %)
    \filldraw[fill=gray!5!white,draw=black] (fsr_ins) -- ($(fsr_ins)+(arcstart_ins)$) arc (90:-90:0.3) -- ($(fsr_ins)+(arcstart_ins)+(arcsep)$)  arc (90:-1:0.3) -- ($(fsr_ins)+(arcstart_ins)+(arcsep)+(arcendr)$) arc (-36.5:54:1) -- cycle;

    \coordinate (Fn_start) at ($(fsl_ins)+(arcsep)+(-0.2,0.4)$);
    \draw[->,very thick,red] (Fn_start) -- ($(Fn_start)+(0,-0.4)$) node[anchor=south east]{$F_a$};
    \draw[->,very thick,red] ($(Fn_start)+(1.4,0)$) -- ($(Fn_start)+(1.4,-0.4)$) node[anchor=south west]{$F_a$};
    
    \draw[->,very thick,blue] (Fn_start) -- ($(Fn_start)+(0.4,0.4)$) node[anchor=south]{$F_n$};
    \draw[->,very thick,blue] ($(Fn_start)+(1.4,0)$) -- ($(Fn_start)+(1,0.4)$) node[anchor=south]{$F_n$};
    
    \draw[->,very thick,black] ($(fsl_ins)+(arcsep)+(0,0.3)$) -- node[anchor=west]{$F_f$} ($(fsl_ins)+(arcsep)+(0,0.5)$);
    \draw[->,very thick,black] ($(fsl_ins)+(arcsep)+(1,0.3)$) -- node[anchor=east]{$F_f$} ($(fsl_ins)+(arcsep)+(1,0.5)$);

    \draw[black,thick] ($(cs_ins)+{1/2}*(cw)+{1/5}*(cl)$) circle (1);
\end{tikzpicture}
    }%
    \subfloat[Force diagram, no interlocking.]{\label{fig:interlock:force_noint}
    \begin{tikzpicture}[scale=1.7]
    \coordinate (cs_ins) at (0,0);
    \coordinate (cw) at (1.4,0);
    \coordinate (cl) at (0,-2);
    \coordinate (cew) at (0.2,0);
    \coordinate (cel) at (0,-0.2);
    \coordinate (csx) at ($2*(cew)$);
    \coordinate (csy) at (cl);
    \filldraw[fill=green!50!black] (cs_ins) -- ($(cs_ins)+(cel)$) -- ($(cs_ins)+(cel)+(cew)$) -- ($(cs_ins)+{0.8}*(csx)+{0.66}*(csy)$) -- ($(cs_ins)+{0.8}*(csx)+{0.66}*(csy)$) arc (247:293:1) -- ($(cs_ins)+(cw)-(cew)+(cel)$) -- ($(cs_ins)+(cw)+(cel)$) -- ($(cs_ins)+(cw)$) -- cycle;

    \coordinate (ins_start) at (0.2,0.4);
    \coordinate (arcstart_ins) at (0,-0.1);
    \coordinate (f_sep) at (1,0);
    \coordinate (fsl_ins) at ($(cs_ins)+(ins_start)$);
    \coordinate (fsr_ins) at ($(cs_ins)+(ins_start)+(f_sep)$);
    \coordinate (arcsep) at (0,-1);
    \coordinate (arcendl) at (-0.3,-0.3);
    \coordinate (arcendr) at (0.3,-0.3);
    \filldraw[fill=gray!5!white,draw=black] (cs_ins) -- ($(cs_ins)+(0,-1.1)$) arc (225:135:1) -- cycle;
    \filldraw[fill=gray!5!white,draw=black] ($(cs_ins)+(cw)$) -- ($(cs_ins)+(cw)+(0,-1.1)$) arc (-45:45:1) -- cycle;

    \coordinate (Fn_start) at ($(cs_ins)$);
    \draw[->,very thick,red] (Fn_start) -- ($(Fn_start)+(0,-0.4)$) node[anchor=south east]{$F_a$};
    \draw[->,very thick,red] ($(Fn_start)+(1.4,0)$) -- ($(Fn_start)+(1.4,-0.4)$) node[anchor=south west]{$F_a$};
    \draw[->,very thick,black] (Fn_start) -- ($(Fn_start)+(0,0.25)$) node[anchor=200]{$F_f$};
    \draw[->,very thick,black] ($(Fn_start)+(1.4,0)$) -- ($(Fn_start)+(1.4,0.25)$) node[anchor=340]{$F_f$};
    
    \draw[->,very thick,blue] (Fn_start) -- ($(Fn_start)+(0.1,0)$) node[anchor=210]{$F_n$};
    \draw[->,very thick,blue] ($(Fn_start)+(1.4,0)$) -- ($(Fn_start)+(1.3,0)$) node[anchor=330]{$F_n$};

    \draw[black,thick] ($(cs_ins)+{1/2}*(cw)+{1/5}*(cl)$) circle (1);
\end{tikzpicture}
    }%
    \vspace{0.1mm}
    \caption{Cannula fitting interlocking with device. With interlocking, applied force ($F_a$) pulling the fitting from the gripper is theoretically resisted by both normal ($F_a$) and shear ($F_f$) forces.}
    \label{fig:interlock}
    \vspace{-3mm}
\end{figure}

Although deflation does not meaningfully change the fingerpads' micro-scale frictional properties, it \textit{does} cause changes in local morphology that enable useful macro-scale grasping behaviour. 
The deflated active regions form recessed areas that can \textit{interlock} with protrusions on certain types of gripped objects. 
To demonstrate this, we applied negative pressure and then gripped a cannula fitting as shown in \cref{fig:interlock:above}. 
The protruding edges of the fitting interlocked with the recessed areas of the fingerpad, increasing cA at the grip interface. 

Interlocking also changes the direction of the theoretical forces that would be experienced by the fitting during shear testing. 
In the neutral case, force applied to the fitting would nominally be resisted by only small shear friction forces (as in \cref{fig:interlock:force_noint}). 
When interlocking, however, the fitting partially contacted internal faces of the active regions, so there would be normal force components directly resisting the applied force (as in \cref{fig:interlock:force}). 

When our deflated fingerpads grasp objects with compatible dimensions, they have the potential to create very strong grasps using interlocking. 
Similarly, inflating the active regions can also create useful macro-scale behaviour. 
Protruding active regions can interlock with small convex objects/object features or partially envelop small concave objects (as in \cref{fig:front}), thereby increasing cA by filling in gaps between the curved object surface and the initially flat fingerpad surface.

\subsection{Demonstration: Pressure Feedback for Object Detection}

As in \cite{PozziEtAlSoftPneumaticPads2024} and \cite{HeEtAlSoftFingertipsTactile2020}, we implement a simple object detection scheme using pressure feedback. 
If a fingerpad is inflated prior to gripping an object, its initial volume $V_0$ decreases to $V<V_0$ when the object contacts and compresses the inflated regions.
Per Boyle's Law, $P_0V_0=PV$ when temperature is constant. 
The initial fingerpad pressure $P_0$ will hence increase to $P>P_0$ on contact with the object. 
We can interpret this sudden increase in pressure as an object detection signal. 

The 2F-85 gripper has a built-in object detection signal, which is set if the current through the motor controlling the gripper fingers exceeds a threshold current level.
We can thus also compare the performance of our pressure-based object detection scheme against that of the 2F-85 object detection functionality.

\begin{figure}[t!]
    \centering
    \begin{tikzpicture}[]   
        \node (plot) at (0,0){\includegraphics[width=0.9\linewidth]{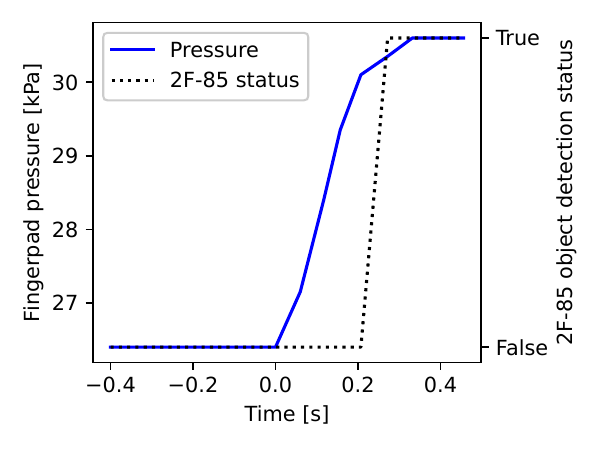}};
        \coordinate (p) at (-1.3,0.1);
        \draw[->,very thick] (p) -- (-0.4,-1.5); 
        \filldraw[very thick,fill=gray!10!white] (p) circle (1.3cm);
        \node (pic) at (p){\includegraphics[width=2.4cm]{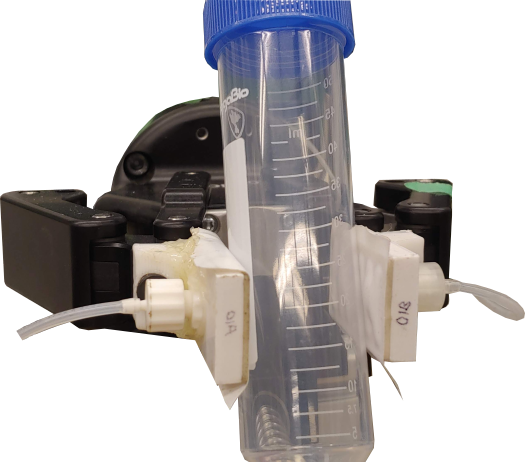}};
        \path ($(p)-(0.6,0.8)$) node[] {A};
    \end{tikzpicture}
    \vspace{-5mm}
    \caption{Change in fingerpad internal pressure and 2F-85 gripper object detection status while gripping a plastic centrifuge tube. The pressure begins to increase well before the 2F-85 detects the gripped object. Inset A shows the gripped tube.}
    \label{fig:demo}
    \vspace{-3mm}
\end{figure}
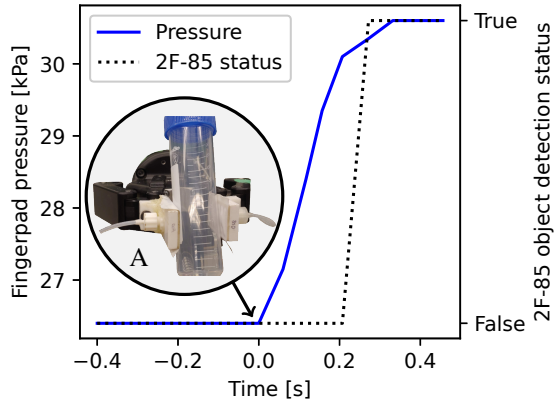

To demonstrate this feature, we inflated our pneumatic fingerpads and then gripped a plastic tube. 
When the fingerpads contacted the tube, their internal pressure readings began to rise immediately. 
The gripper's object detection flag activated after more than 0.2 seconds, by which time the fingerpad pressure had already rapidly risen as shown in \cref{fig:demo}. 
This demonstration shows that our fingerpads enable simple pressure-based object detection that is faster than the 2F-85 gripper's built-in object detection. 

\subsection{Limitations}

One limitation of our design is its reliance on pneumatic control. 
As with any pneumatically actuated soft surface, our fingerpads are vulnerable to punctures and leaks. 
The fingerpads also require air to be supplied via inlet tubing, which may impose constraints on their compatibility with grippers that twist or have unavoidable pinch points. 
In addition, pneumatic control is energetically expensive, particularly if there are minor leaks in the fingerpad or inlet tubing.

While the continuous top layer used in our design improves robustness compared to designs with multiple moving layers (e.g., \cite{BeckerEtAlTunableFrictionConstrained2017}) by reducing the risk of snagging, using a single contact layer also limits friction tuning.  
Since actuation only changes the morphology of the contact surface rather than completely swapping contact surfaces, our range of friction force variance is lower than that of designs that use separate high-friction and low-friction contact surfaces. 

Additionally, only a small set of objects was used during shear force testing. 
While these initial results are promising, shear force tests should be repeated with objects that have more complex shapes, larger masses, and different surface textures. 
The fingerpads should also be implemented in and tested with a larger selection of gripper designs to assess the impact of changes in finger count and configuration on the frictional properties of the fingerpads.

\section{Conclusion}
\label{sec:Conclusion}

In this paper, we have introduced structured pneumatic fingerpads that use local morphology changes to tune friction at the grip interface. 
We have characterized the fingerpads' frictional behaviour and shown that friction forces at the grip interface can be adjusted by up to a factor of 2.8. 
We have also shown examples in which the fingerpads enable different modes of gripping such as interlocking and enveloping. 
Finally, we have demonstrated that our fingerpads can be used to detect object contact using pressure feedback. 

In comparison to existing solutions, our design is significantly easier to fabricate and repair. 
A major strength of this design is the separation of roles between the layers: the base layer provides both pneumatic inputs and rigidity, the middle layer defines active regions, and the top layer acts as a contact surface. 
Each individual layer can be easily customized or replaced with only minimal changes to other layers. 
Moreover, the layers are fabricated using common prototyping materials and methods, and the total material cost per fingerpad is less than 1 USD. 
Most importantly, our design can be easily implemented in different industrial grippers by modifying the base layer and its connector.

The PVC-covered prototype characterized in this work serves as a proof of concept for the overall structured pneumatic fingerpad design. 
In future work, we will fabricate and characterize devices using other contact surfaces. 
In particular, we intend to use resin printing to fabricate flexible top layers with custom microtopography in the active regions.

\bibliographystyle{ieeetr}
\bibliography{refs}

\end{document}